\pdfoutput=1

\documentclass[11pt]{article}

\usepackage{emnlp2021}

\usepackage{times}
\usepackage{latexsym}
\usepackage{graphics}
\usepackage{graphicx}
\usepackage[figuresright]{rotating}
\usepackage{multirow}
\usepackage{makecell}
\usepackage{lscape}
\usepackage{setspace}
\usepackage{amssymb}
\usepackage[T1]{fontenc}

\usepackage[utf8]{inputenc}

\usepackage{microtype}

\title{Integrating knowledge bases to improve coreference and bridging resolution for the chemical domain}

\author{Pengcheng Lu \and
  Massimo Poesio\\
School of Electronical Engineering and
Computer Science\\Queen Mary University of London, United Kingdom\\
  \texttt{\{pengcheng.lu, m.poesio\}@qmul.ac.uk
  }
}
\begin{document}
\maketitle
\begin{abstract}
Resolving coreference and bridging relations in chemical patents is important for better understanding the precise chemical process, where chemical domain knowledge is very critical. We proposed an approach incorporating external knowledge into a multi-task learning model for both coreference and bridging resolution in the chemical domain. The results show that integrating external knowledge can benefit both chemical coreference and bridging resolution.

\end{abstract}

\section{Introduction}

Due to the large volume of chemical patents and literature, manually chemical information extraction is costly, which can be greatly improved by developing automated natural language processing systems for chemical documents \citep{krallinger2015chemdner}. Chemical patents contain rich coreference and bridging links, such as chemical reaction process. Resolving these links can help us understand these precise chemical processes.

Coreference resolution is a fundamental and key component of many downstream natural language processing tasks, such as information extraction. Compared with coreference, bridging resolution is more difficult and less studied \cite{poesio2023computational}. \citet{yu2020multitask} and \citet{fang2021chemu} have shown that coreference resolution and bridging resolution can benefit each other via a joint neural model. Similar with them, we adopt a multi-task end-to-end architecture jointly predicting coreference and bridging resolution for the chemical domain. 

But different from them, we integrate external knowledge base into the end-to-end model. Besides linguistic and contextual information, chemical domain knowledge is crucial for correctly resolving chemical coreference and bridging links. We align spans to corresponding entities from external knowledge base, and then combine span and entity embedding to generate a knowledge-enriched span embedding for better resolving coreference and bridging in the chemical domain. Our proposed model achieved better performance on both chemical coreference and bridging resolution.

\section{Related Work}
\label{Related Work}
The datasets and models for chemical coreference and bridging are still less studied. \citet{fang2021chemu} annotated ChEMU-REF dataset with both coreference and bridging resolution labels, consisting of reaction description snippets from chemical patents. They adopted an end-to-end architecture with multi-task learning for resolving coreference and bridging. \citet{machi2021hukb} used a two-step approach on ChEMU-REF, where mentions including both antecedent and anaphor are first detected, then mentions are classified into different types and relations between them are determined. \citet{dutt2021pipelined} also proposed a pipelined system combining learning-based and rule-based methods and achieved better performance on ChEMU-REF.

Our proposed model is based on \citet{fang2021chemu}, but the differences from them are: (1) they used LSTM as the encoder with GloVe, character and ELMo embeddings as the inputs, while we used SpanBERT as the encoder, which has been proven its powerful ability on span-related tasks. (2) they used ChEMLo embeddings pre-trained on the chemical domain to integrate domain knowledge, while we incorporated external knowledge base into the end-to-end model, which encoded rich structural information beyond contextual information. (3) to improve the computation efficiency, we used chemical tokenizer to reduce the number of candidate spans.

\section{Method}
\label{method}

\begin{figure*}[h]
\centering
\includegraphics[width=1.82\columnwidth]{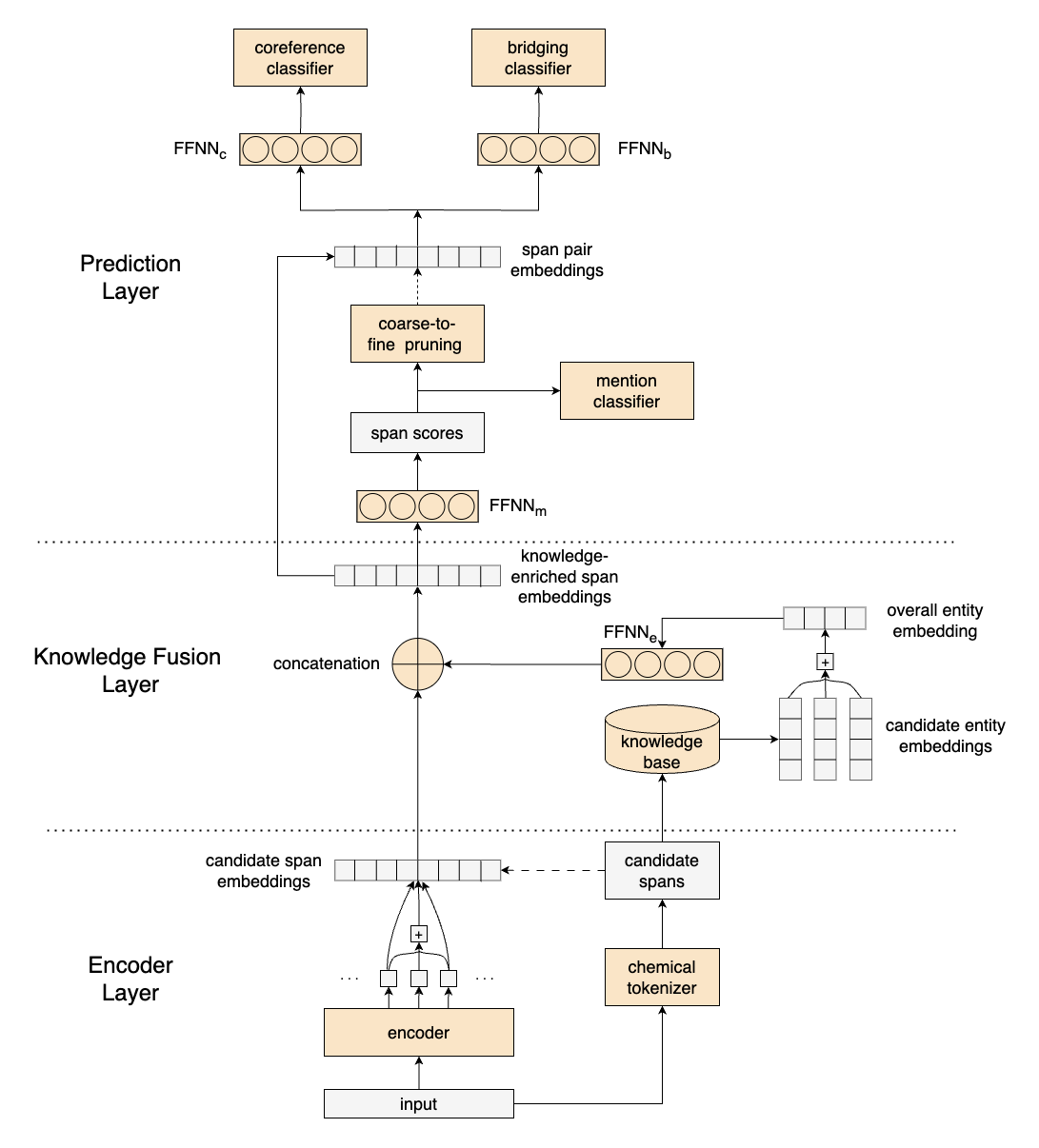}
\caption{The framework of our proposed model for chemical coreference and bridging resolution}
\label{chemical_model}
\end{figure*}

In this section, we present the overall framework of our proposed model and the detailed implementation. As shown in Figure \ref{chemical_model}, the framework contains three layers: (1) encoder layer, (2) knowledge fusion layer and (3) prediction layer. First we input documents into the encoder and produce representations for candidate spans. Then, an entity linking tool is used to align candidate spans to their corresponding entities in the knowledge base. Knowledge-enriched span embeddings are generated by combining span embeddings and entity embeddings. Finally, in the prediction layer, the knowledge-enriched embeddings are taken as the input to predict the coreference and bridging relations for candidate spans. 
 
\subsection{Encoder Layer}
We choose SpanBERT \citep{joshi2020spanbert} as the encoder to generate representations for candidate spans, which can better learn span-level information and has achieved great performance on a variety of tasks especially span selection tasks, such as question answering and coreference resolution. Given an input document, SpanBERT outputs the contextualized representation for each input token.

Following \citet{lee2018higher}, for each span $i$, its representation is computed as follows:
\begin{equation}
s_{i}=[x_{STARTi},x_{ENDi},h_{i}, \phi_{i}]
\end{equation}
where $x_{STARTi}$, $x_{ENDi}$ are the start and end token representation of span $i$, and $\phi_{i}$ is a feature vector encoding the width of span $i$. And $h_{i}$ is a weighted sum of token representation in the span $i$, which is computed using an attention mechanism as follows:
\begin{equation}
\alpha_{t}=\emph{FFNN}_{\alpha}(x_{t})
\end{equation}
\begin{equation}
a_{i,t}=\frac{e^{\alpha_{t}}}{\sum_{j=START(i)}^{END(i)}e^{\alpha_{j}}}
\end{equation}
\begin{equation}
h_{i}=\sum_{t=START(i)}^{END(i)}a_{i,t}x_{t}
\end{equation}
where ${FFNN}_{\alpha}$ is a feed-forward neural network mapping from each token representation $x_{t}$ to its unnormalized attention score $\alpha_{t}$, and $a_{i,t}$ is the normalized attention score of token $t$ in the span $i$.

In \citet{lee2018higher}, they consider all possible spans as the candidate spans. However, this approach does not work well in chemical domain when using BERT tokenizer. The chemical names in the text are often long and contain numbers and symbols, such as "-", "(" and ")". Hence a chemical name could be broken up into lots of tokens, which will greatly increase the number of candidate spans if considering all possible spans. Especially, the entity linking tool, which is very time consuming, is used in the following steps of our proposed model. As the number of candidate spans increases, the computation efficiency will be greatly reduced. Therefore, we use OSCAR4 \cite{jessop2011oscar4}, which is an open source chemical text-mining toolkit, to tokenize the input document and generate candidate spans based on the output of OSCAR4 tokenizer. Then the candidate spans are used to look up the corresponding candidate span embeddings from the output of the encoder.

\subsection{Knowledge Fusion Layer}
Knowledge fusion layer aims at obtaining corresponding entities from the knowledge base according to candidate spans and producing final knowledge-enriched span representations by combining span and entity embeddings. There are three steps: (1) entity linking; (2) knowledge representation; and (3) span-entity fusion.

First, the entity linking tool scispaCy \citep{neumann2019scispacy} is used to link the candidate spans to corresponding entities in the knowledge base UMLS\footnote{\url{https://www.nlm.nih.gov/research/umls/index.html}}. scispaCy uses an approximate nearest neighbours search based on the vector of TF-IDF scores of character 3-grams.

After performing entity linking, the top K candidate entities for each span are retrieved. Then, for each candidate entity we look up the corresponding pre-trained entity embeddings from \citet{maldonado2019adversarial}, which is an extension of KBGAN adversarial framework \cite{cai2018kbgan}, using Generative Adversarial Networks (GANs) for learning embeddings representing entities, relations between them, semantic types and semantic relations in UMLS. After obtaining the entity embeddings of candidate entities, we generate the overall entity embedding by weighted averaging the candidate entity embeddings which are linked to the same span:

\begin{equation}
e_{i}=\sum_{k}w_{ik}e_{ik}
\end{equation}
where $e_{ik}$ is the candidate entity embedding linked to span $i$, $w_{ik}$ is the similarity score between the candidate entity $e_{ik}$ and the span $i$ based on TF-IDF vectors from scispaCy, and $e_{i}$ is the overall entity embedding for span $i$.

 Due to the coverage of the knowledge base UMLS on the chemical domain, there are always some cases that the entity linker can not find entities in the knowledge base for spans. When the corresponding entities are absent, we use zero vectors of the same dimensions to resolve these cases.

Finally, the overall entity embeddings are taken as the input of a feed-forward neural network, which aims at converting the entity embeddings to a new vector space for a better fusion with span embedding. The outputs are concatenated with the corresponding span embeddings to generate the knowledge-enriched span embeddings:

\begin{equation}
g_{i}=[s_{i}, \emph{FFNN}_{e}(e_{i})]
\end{equation}
where $s_{i}$ is the span embedding, $e_{i}$ is the entity embedding, and $g_{i}$ is the knowledge-enriched span embedding.

\subsection{Prediction Layer}
Following \citet{fang2021chemu} and \citet{yu2020multitask}, the prediction layer adopts a multi-task architecture including mention detection, coreference resolution and bridging resolution. For coreference and bridging resolution, the goal is to learn a distribution $P (y_{i}$) over possible antecedents $Y(i)$ for each span $i$:
\begin{equation}
P(y_{i})=\frac{e^{s(i,y_{i})}}{\sum_{y^{'}\in Y(i)}e^{s(i,y^{'})}}
\end{equation}
where $s(i, j)$ is a pairwise score for a coreference or bridging link (denoted as $s_c(i, j)$ and $s_b(i, j)$ respectively) between span $i$ and span $j$. The pairwise score is computed as follows:
\begin{equation}
s_{c}(i,j)=\left\{
\begin{array}{llllllllll}
	0                          & j=\epsilon   \\
	\emph{FFNN}_c\left(g_{i,j}\right) & j\ne\epsilon
\end{array}\right.
\end{equation}
\begin{equation}
s_{b}(i,j)=\emph{FFNN}_{b}(g_{i,j})
\end{equation}
\begin{equation}
g_{i,j}=[g_{i},g_{j},g_{i}\circ g_{j},\phi (i,j)]
\end{equation}
where $g_{i,j}$ is the span pair embedding, $g_{i}$ and $g_{j}$ are knowledge-enriched span embeddings for span $i$ and $j$, ◦ denotes element-wise product, $\phi (i,j)$ represents the distance feature between a span pair, and $\epsilon$ represents a dummy antecedent to deal with the case that the span is either not a mention, or is not coreferent with any previous span.

To maintain computation efficiency, we only consider up to $\lambda T$ spans with the highest span scores, where $T$ is the document length. The span scores are computed as follows:
\begin{equation}
s_{m}(i)=\emph{FFNN}_{m}(g_{i})
\end{equation}

Then, the coarse-to-fine antecedent pruning is further conducted to keep the top K antecedents of each remaining span, as \citet{lee2018higher}.

\begin{table}
\centering
\resizebox{\linewidth}{!}{
\begin{tabular}{lllll}
\hline
\textbf{} & \textbf{Train} & \textbf{Dev} & \textbf{Test} & \textbf{Total}\\
\hline
snippets & 900 & 225 & 375 & 1500\\
\hline
coreference & 2624 & 619 & 1491 & 4734\\
bridging & 10377 & 2419 & 4135 & 16931\\
\hline
transformed & 493 & 107 & 166 & 766\\
reaction-associated & 3308 & 764 & 1245 & 5317\\
work-up & 6230 & 1479 & 2576 & 10285\\
contained & 346 & 69 & 148 & 563\\
\hline
\end{tabular}
}
\caption{\label{statistics_Chem}
Statistics of ChEMU-REF.  “bridging” is the total
across all bridging relations
}
\end{table}

\begin{table*}
\centering
\resizebox{\linewidth}{!}{
\begin{tabular}{llllllll}
\hline
\multirow{2}{*}{\textbf{Relation type}} & \multirow{2}{*}{\textbf{Model}} & \multicolumn{3}{c}{\textbf{Mention (anaphor)}} & \multicolumn{3}{c}{\textbf{Relation}}\\
\cline{3-8}
&  & \textbf{P} & \textbf{R} & \textbf{F1} & \textbf{P} & \textbf{R} & \textbf{F1}\\
\hline
\multirow{4}{*}{Coreference} & ChemBERT+UMLS & 90.61 & 55.41 & 68.76 & 82.39 & 42.05 & 55.68\\
& SciBERT+UMLS  & 91.97 & 50.43 & 65.14 & 82.92 & 38.43 & 52.52\\
& BERT+UMLS & 91.49 & 56.56 & 69.90 & 83.27 & 45.07 & 58.49\\
& SpanBERT+UMLS & \textbf{92.14} & \textbf{61.72} & \textbf{73.93} & \textbf{83.50} & \textbf{50.23} & \textbf{62.73} \\
\hline
\multirow{4}{*}{Bridging} & ChemBERT+UMLS & 88.19 & 73.17 & 79.98 & 62.90 & 61.60 & 62.24\\
& SciBERT+UMLS & 90.04 & 80.82 & 85.18 & 80.40 & 68.85 & 74.18\\
& BERT+UMLS & 91.87 & 82.91 & 87.16 & \textbf{84.55} & 74.90 & 79.43\\
& SpanBERT+UMLS & \textbf{91.89} & \textbf{86.18} & \textbf{88.94} & 83.23 & \textbf{79.56} & \textbf{81.36} \\
\hline
\multirow{4}{*}{Overall} & ChemBERT+UMLS & 88.89 & 66.79 & 76.27 & 65.99 & 56.42 & 60.83\\
& SciBERT+UMLS & 90.53 & 69.92 & 78.90 & 80.81 & 60.79 & 69.39\\
& BERT+UMLS & 91.76 & 73.45 & 81.59 & \textbf{84.32} & 66.99 & 74.66\\
& SpanBERT+UMLS & \textbf{91.96} & \textbf{77.40} & \textbf{84.06} & 83.28 & \textbf{71.79} & \textbf{77.11} \\
\hline
\end{tabular}
}
\caption{\label{result}
Anaphor detection and relation prediction results on the test set(\%). P and R are short for Precision and Recall.
}
\end{table*}

\begin{table*}
\centering
\resizebox{\linewidth}{!}{
\begin{tabular}{llllllll}
\hline
\multirow{2}{*}{\textbf{Relation type}} & \multirow{2}{*}{\textbf{Model}} & \multicolumn{3}{c}{\textbf{Strict Matching}} & \multicolumn{3}{c}{\textbf{Relaxed Matching}}\\
\cline{3-8}
&  & \textbf{P} & \textbf{R} & \textbf{F1} & \textbf{P} & \textbf{R} & \textbf{F1}\\
\hline
\multirow{4}{*}{Coreference} & Baseline-ELMo & \textbf{84.97} & 44.74 & 58.61 & 91.85 & 48.36 & 63.36 \\
& Baseline-ChELMo & 84.76 & 46.61 & 60.15 & \textbf{92.44} & 50.84 & 65.60 \\
& SpanBERT & 83.06 & 48.29 & 61.07 & 89.73 & 52.18 & 65.99 \\
& SpanBERT+UMLS & 83.50 & \textbf{50.23} & \textbf{62.73} & 90.19 & \textbf{54.26} & \textbf{67.76} \\
\hline
\multirow{4}{*}{Contained} & Baseline-ELMo &  91.75 & 60.14 & 72.65 & \textbf{97.94} & 64.19 & 77.55 \\
& Baseline-ChELMo & \textbf{92.11} & 70.95 & \textbf{80.15} & 93.86 & 72.30 & \textbf{81.68} \\
& SpanBERT & 67.46 & \textbf{77.03} & 71.92 & 70.41 & \textbf{80.41} & 75.08 \\
& SpanBERT+UMLS & 80.45 & 72.30 & 76.16 & 84.21 & 75.68 & 79.72 \\
\hline
\multirow{4}{*}{Reaction Associated} & Baseline-ELMo &  81.45 & 72.29 & 76.60 & 84.98 & 75.42 & 79.91\\
& Baseline-ChELMo & \textbf{83.81} & 73.57 & 78.36 & \textbf{86.73} & 76.14 & 81.09 \\
& SpanBERT & 77.20 & 76.95 & 77.07 & 80.98 & 80.72 & 80.85 \\
& SpanBERT+UMLS & 81.96 & \textbf{77.75} & \textbf{79.80} & 85.35 & \textbf{80.96} & \textbf{83.10} \\
\hline
\multirow{4}{*}{Transformed} & Baseline-ELMo & 78.77 & 84.94 & 81.74 & 78.77 & 84.94 & 81.74\\
& Baseline-ChELMo & 79.35 & 87.95 & 83.43 & 79.35 & 87.95 & 83.43 \\
& SpanBERT & 78.17 & \textbf{92.77} & 84.85 & 78.17 & \textbf{92.77} & 84.85 \\
& SpanBERT+UMLS & \textbf{81.48} & \textbf{92.77} & \textbf{86.76} & \textbf{81.48} & \textbf{92.77} & \textbf{86.76} \\
\hline
\multirow{4}{*}{Work Up} & Baseline-ELMo & 85.66 & 76.05 & 80.57 & 89.90 & 79.81 & 84.56 \\
& Baseline-ChELMo & \textbf{87.05} & 78.03 & \textbf{82.29} & \textbf{91.81} & 82.30 & 86.80 \\
& SpanBERT & 82.33 & 79.04 & 80.65 & 86.86 & 83.39 & 85.09 \\
& SpanBERT+UMLS & 84.12 & \textbf{80.01} & 82.01 & 89.31 & \textbf{84.94} & \textbf{87.07} \\
\hline
\multirow{4}{*}{Overall} & Baseline-ELMo & 84.35 & 66.76 & 74.53 & 88.75 & 70.25 & 78.42\\
& Baseline-ChELMo & \textbf{85.66} & 68.82 & 76.33 & \textbf{90.24} & 72.50 & 80.41 \\
& SpanBERT & 80.49 & 70.78 & 75.32 & 84.98 & 74.72 & 79.52 \\
& SpanBERT+UMLS & 83.28 & \textbf{71.79} & \textbf{77.11} & 88.06 & \textbf{75.92} & \textbf{81.54} \\
\hline
\end{tabular}
}
\caption{\label{comparison_results}
Performance per relation type compared with baselines on the test set.(\%). P and R are short for Precision and Recall.
}
\end{table*}

\section{Experiments}
\label{experiment}
\subsection{Experimental Setup}
\paragraph{Datasets}
 The experiments are conducted on ChEMU-REF 2021 dataset, which is a part of ChEMU 2021 shared task, consisting of reaction description snippets extracted from chemical patents. The details of ChEMU-REF 2021 are shown in Table \ref{statistics_Chem}. The dataset contains both coreference and bridging resolution annotations. In particular, four sub-types of bridging relations are defined, which are Transformed, Reaction-associated, Work-up and Contained respectively.

\paragraph{Knowledge Base} 
We used Unified Medical Language Systems (UMLS) as the knowledge resource. UMLS is the largest biomedical KB, consisting of metathesaurus, semantic network, and lexical resources. Metathesaurus is the major component of UMLS, containing concepts (or called entities) from nearly 200 different source vocabularies and different types of relations connecting these concepts. UMLS also contains chemical entities besides biomedical. The version of UMLS which we used is the newest 2023 AB release.

\paragraph{Implementation Details}
We use ChemBERT \cite{guo2021automated}, SciBERT \cite{beltagy2019scibert}, BERT\_base \cite{devlin2018bert} and SpanBERT\_base \cite{joshi2020spanbert} as the encoder respectively. These models are fine-tuned on ChEMU-REF for 20 epochs, using learning rate of $1\times10^{-5}$ for PLMs parameters and $2\times10^{-4}$ for task parameters with Adam optimizer, a dropout of 0.3, \verb|max_training_len| of 384, and \verb|max_span_width| of 120. For other hyperparameters, we use the default settings from \citet{lee2018higher}.

\paragraph{Evaluation}
For the evaluation, we used the BRATEval script provided by the ChEMU-REF organizers. The performances are evaluated in terms of precision, recall, and F1 score, for both exact matching and relaxed matching.

\subsection{Results}

The precision, recall and F1 scores of our model on mention detection and relation prediction are shown in Table \ref{result}. Note that, the evaluation for mention detection just considers anaphors. In the experiment, different encoders are used and combined with UMLS, including ChemBERT, SciBERT, BERT\_base and SpanBERT\_base. The comparison results show that SpanBERT achieved the best performance on both coreference and bridging resolution, which shows the powerful ability of SpanBERT on chemical coreference and bridging even without pre-training on chemical domain.  

The performance per relation types compared with the baselines are shown in Table \ref{comparison_results}, for both strict and relaxed matching. The results of \textit{Baseline-ELMo} and \textit{Baseline-ChELMo} are from \citet{li2021extended}, and \textit{SpanBERT} denotes our proposed model without integrating the knowledge base as a comparison. The comparison results show that our proposed model integrating the knowledge base is uniformly superior to the model without the knowledge base, which shows that integrating external knowledge base can benefit both coreference and bridging resolution on chemical domain. Compared with the baseline models, our proposed model achieved better performance for strict matching on difference relation types except \textit{Contained} and \textit{Work Up}. Especially, our proposed model generally achieved higher recall but lower precision compared with the baseline models. 

\begin{table}
\centering
\resizebox{\linewidth}{!}{
\begin{tabular}{llll}
\hline
\textbf{Relation Type} & \textbf{\#Mention} & \textbf{\#Disc.} & \textbf{Proportion of Disc.}\\
\hline
coreference & 3552 & 615 & 17.31\%\\
bridging & 8460 & 731 & 8.64\%\\
\hline
transformed & 611 & 4 & 0.65\%\\
reaction-associated & 3018 & 305 & 10.11\%\\
work-up & 4201 & 402 & 9.57\%\\
contained & 630 & 20 & 3.17\%\\
\hline
\end{tabular}
}
\caption{\label{disc. mention}
Statistics of discontinuous mentions in ChEMU-REF.  “bridging” is the total across all bridging relations, and \textit{\#Disc.} means the number of discontinuous mentions
}
\end{table}

\begin{table*}
\centering
\resizebox{\textwidth}{!}{
\begin{tabular}{lllllllllll}
\hline
\textbf{Proportion} & \multicolumn{5}{c}{\textbf{subtoken length}} & \multicolumn{5}{c}{\textbf{token length}}\\
\hline
 & Coref. & TR & RA & WU & CT & Coref. & TR & RA & WU & CT\\
\hline
L<=10 & 55.65 & 97.97 & 62.64 & 85.24 & 51.75 & 93.38 & 98.56 & 88.97 & 96.55 & 90.41\\
10<L<=30 & 21.83 & 0.66 & 26.93 & 11.44 & 29.98 & 6.62 & 0.79 & 10.64 & 3.42 & 7.56\\
30<L<=60 & 11.09 & 0.26 & 6.68 & 1.74 & 12.82 & 0 & 0.59 & 0.39 & 0.03 & 2.03 \\
60<L<=90 & 7.25 & 0.20 & 2.18 & 1.07 & 3.32 & 0 & 0.07 & 0 & 0 & 0 \\
90<L<=120 & 2.97 & 0.66 & 1.23 & 0.31 & 1.38 & 0 & 0 & 0 & 0 & 0 \\
L>120 & 1.22 & 0.26 & 0.33 & 0.19 & 0.74 & 0 & 0 & 0 & 0 & 0 \\
\hline
average length & 21.05 & 3.99 & 13.05 & 7.02 & 17.89 & 4.16 & 2.99 & 5.56 & 3.78 & 6.50 \\
\hline
\end{tabular}
}
\caption{\label{proportion_length}
proportion of different length of mentions by different relation types
}
\end{table*}

\begin{table*}
\centering
\resizebox{\textwidth}{!}{
\begin{tabular}{lllllllllll}
\hline
\textbf{Proportion} & \multicolumn{5}{c}{\textbf{subtoken distance}} & \multicolumn{5}{c}{\textbf{token distance}}\\
\hline
 & Coref. & TR & RA & WU & CT & Coref. & TR & RA & WU & CT\\
\hline
D<=30 & 50.65 & 88.06 & 55.19 & 88.13 & 74.91 & 55.94 & 95.93 & 79.18 & 96.65 & 91.51\\
30<D<=60 & 7.70 & 9.84 & 29.24 & 11.11 & 17.53 & 10.10 & 3.94 & 18.76 & 3.08 & 8.30\\
60<D<=90 & 4.78 & 1.57 & 10.21 & 0.54 & 5.35 & 7.19 & 0 & 1.86 & 0.18 & 0.18\\
90<D<=120 & 3.67 & 0.39 & 3.46 & 0.14 & 1.85 & 10.86 & 0.13 & 0.12 & 0.06 & 0\\
D>120 & 33.20 & 0.13 & 1.90 & 0.09 & 0.37 & 15.91 & 0 & 0.08 & 0.02 & 0\\
\hline
average distance & 90.33 & 18.79 & 33.22 & 15.99 & 19.89 & 51.58 & 15.36 & 19.80 & 12.07 & 11.86 \\
\hline
\end{tabular}
}
\caption{\label{proportion_distance}
proportion of different distance between anaphors and antecedents by different relation types
}
\end{table*}

\section{Discussion}
\subsection{coreference vs. bridging}
Generally, bridging resolution is a more challenging task than coreference. However, as shown in Table \ref{comparison_results}, the performances of our models on different types of bridging relations are much better than coreference for ChEMU-REF. By looking into the statistics and details of the dataset, the reasons may be as follows: 

(1) The number of bridging relations annotated in ChEMU-REF is much larger than coreference (78\% vs. 22\%), as shown in Table \ref{statistics_Chem}; 

(2) The proportion of discontinuous mention in coreference is almost twice as much as bridging (17.31\% vs. 8.64\%) as shown in Table \ref{disc. mention}. However, our proposed model can not deal with discontinuous mentions, which means the recall of our models on coreference has been reduced by 17.31\% due to the discontinuous mentions; 

(3) We tokenized the dataset using both BERT tokenizer and chemical tokenizer. A long chemical name will be broken up into lots of subtokens by BERT tokenizer, while the chemical tokenizer (such as OSCAR4) can keep the long chemical name as one token as discussed before. As shown in Table \ref{proportion_length}, the average length of coreference mentions is increased from 4.16 for tokens to 21.05 for subtokens. This indicates that coreference mentions contain lots of long chemical names, which are more difficult to be learnt by our models. We can see the order of F1 scores of our model with strict matching in Table \ref{comparison_results} is just the opposite of the order of the average subtoken length by relation types in Table \ref{proportion_length}, i.e., the shorter the average subtoken length, the higher the F1 score;  

(4) The distances between anaphor and antecedence by relations types are shown in Table \ref{proportion_distance}. We can see the average subtoken distance for coreference is much larger than other relations, and the distance over 120 for coreference still accounts for 33.2\%. We did find quite a few cases of coreference relations in the dataset that the antecedent and the anaphor are at the beginning and the end of the document respectively, which are difficult to resolve by our models.

\begin{table*}
\centering
\resizebox{\linewidth}{!}{
\begin{tabular}{llllllll}
\hline
\multirow{2}{*}{\textbf{Relation type}} & \multirow{2}{*}{\textbf{Training method}} & \multicolumn{3}{c}{\textbf{Mention (anaphor)}} & \multicolumn{3}{c}{\textbf{Relation}}\\
\cline{3-8}
&  & \textbf{P} & \textbf{R} & \textbf{F1} & \textbf{P} & \textbf{R} & \textbf{F1}\\
\hline
\multirow{2}{*}{Coreference} & coreference & 88.78 & 60.24 & 71.77 & 83.44 & 46.32 & 59.57\\
& multi-task & \textbf{92.14} & \textbf{61.72} & \textbf{73.93} & \textbf{83.50} & \textbf{50.23} & \textbf{62.73} \\
\multirow{2}{*}{Bridging} & bridging & 90.68 & 85.97 & 88.26 & 80.65 & 78.91 & 79.77 \\
& multi-task & \textbf{91.89} & \textbf{86.18} & \textbf{88.94} & \textbf{83.23} & \textbf{79.56} & \textbf{81.36} \\
\hline
\multirow{2}{*}{Contained} & bridging & \textbf{95.24} & 81.08 & \textbf{87.59} & 70.39 & \textbf{72.30} & 71.33\\
& multi-task & 92.42 & \textbf{82.43} & 87.14 & \textbf{80.45} & \textbf{72.30} & \textbf{76.16} \\
\multirow{2}{*}{Reaction Associated} & bridging & 90.67 & 87.82 & 89.22 & 80.5 & 77.27 & 78.85\\
& multi-task & \textbf{92.14} & \textbf{88.66} & \textbf{90.36} & \textbf{81.96} & \textbf{77.75} & \textbf{79.80} \\
\multirow{2}{*}{Transformed} & bridging & \textbf{84.92} & 91.57 & \textbf{88.12} & 81.28 & 91.57 & 86.12\\
& multi-task& 83.7 & \textbf{92.77} & 88.00 & \textbf{81.48} & \textbf{92.77} & \textbf{86.76} \\
\multirow{2}{*}{Work Up} & bridging & 91.38 & \textbf{84.71} & 87.92 & 81.29 & 79.27 & 80.27\\
& multi-task & \textbf{93.19} & 84.45 & \textbf{88.61} & \textbf{84.12} & \textbf{80.01} & \textbf{82.01} \\
\hline
\end{tabular}
}
\caption{\label{multitask_results}
Anaphor detection and relation prediction results on the test set(\%). Models are trained for “coreference”, “bridging” or “multi-task”. P and R are short for Precision and Recall.
}
\end{table*}

\subsection{multi-task vs. single-task training}
In addition to the multi-task training, we also conduct the single-task training where our models are only trained on coreference task or bridging task as the comparison, as shown in Table \ref{multitask_results}. The experimental results show that our models via multi-task training outperform the models via only coreference or bridging training, which proves that coreference and bridging can be tackled together and benefit each other through multi-task training.
\subsection{chemical tokenization}
The main purpose of chemical tokenization in our model is to improve computation efficiency by reducing the number of candidate spans through avoiding splitting long chemical names. We find the number of candidate spans can be reduced to one in eight by using chemical tokenization, and the time required for entity linking processing one document is reduced from around 410s to 50s. As regards to the performance, it can maintain the recall and improve the precision a little (around 1\%) compared with models without using chemical tokenizer.

\section{Conclusion}
\label{conclusion}
In this paper we proposed a model integrating external knowledge base into a multi-task learning architecture and achieved better performance on both chemical coreference and bridging. This shows that integrating external knowledge base is an effective way to improve both coreference and bridging resolution for the chemical domain.

\section*{Acknowledgements}
This research was supported in part by the China Scholarship Council, and the ARCIDUCA project, EPSRCEP/W001632/1.

\bibliographystyle{acl_natbib}

\end{document}